\documentclass[fleqn,10pt]{wlscirep}
\usepackage[utf8]{inputenc}
\usepackage[T1]{fontenc}
\usepackage{amsmath,amssymb}
\DeclareMathOperator{\Tr}{Tr}

\usepackage[lined,ruled]{algorithm2e}

\newcommand{\innerR}{R_I}
\newcommand{\outerR}{R_O}
\newcommand{\innerN}{N_I}
\newcommand{\outerN}{N_O}

\title{Neuronal diversity can improve machine learning for physics and beyond}

\author[1,2]{Anshul Choudhary}
\author[1]{Anil Radhakrishnan}
\author[1,3,*]{John F. Lindner}
\author[4]{Sudeshna Sinha}
\author[1]{William L. Ditto}

\affil[1]{Nonlinear Artificial Intelligence Laboratory, Physics Department, North Carolina State University, Raleigh, NC 27607, USA }
\affil[2]{The Jackson Laboratory for Genomic Medicine, Farmington, CT 06032, USA}
\affil[3]{Physics Department, The College of Wooster, Wooster, OH 44691, USA}
\affil[4]{Indian Institute of Science Education and Research Mohali, Knowledge City, SAS Nagar, Sector 81, Manauli PO 140 306, Punjab,
India}

\affil[*]{jlindner@wooster.edu}

\begin{abstract}
    Diversity conveys advantages in nature, yet homogeneous neurons typically comprise the layers of artificial neural networks. Here we construct neural networks from neurons that learn their own activation functions, quickly diversify, and subsequently outperform their homogeneous counterparts on image classification and nonlinear regression tasks. Sub-networks instantiate the neurons, which meta-learn especially efficient sets of nonlinear responses. Examples include conventional neural networks classifying digits and forecasting a van der Pol oscillator and physics-informed Hamiltonian neural networks learning H\'enon-Heiles stellar orbits and the swing of a video recorded pendulum clock. Such \textit{learned diversity} provides examples of dynamical systems selecting diversity over uniformity and elucidates the role of diversity in natural and artificial systems.
\end{abstract}

\begin{document}

\flushbottom
\maketitle 
\thispagestyle{empty}

\section*{Introduction}

Diversity is a hallmark of many complex systems in physics~\cite{Anderson1972,Bak1987}
and in \textit{physics beyond physics}~\cite{Holovatch_2017}, including  microscopic cell populations~\cite{neural_div_cell}, marine and terrestrial ecosystems~\cite{diversity_ecosystem, choudhary2021weak}, financial markets \cite{may_economics}, and social networks~\cite{page2010diversity, stability_and_complexity,may_wigner_generality}. In particular,  mammalian brains contain billions of neurons with diverse cell types whose complex dynamical patterns are believed responsible for the rich range of cognition, affect, and behavior~\cite{marcus2014atoms,neural_div_brain_activity, computational_div, intermediate_div}. But despite the widespread appreciation of diversity in neuroscience, researchers have just begun to explore the role of diversity and adaptability in artificial neural networks~\cite{heterogeneous_cell_types, neural_heterogeneity_natcomm, Han2021}. 

Inspired by nature, artificial neural networks are nonlinear systems that can be trained to learn, classify, and predict. Conventional artificial neural networks contain identical neurons in each network layer, even if the neurons vary from layer to layer. But uniform neuronal activation functions can reduce expressiveness and adaptability, limiting the neural network's capacity to capture the rich diversity of computation and interaction observed in nature. Diversifying the activation functions can overcome such limitations, enabling the networks to be more expressive and better represent the complexity of natural systems. In this article, we propose a novel way to diversify a neural network by learning the neuron types \textit{within} each layer. We flexibly realize the different neurons using sub-networks, or networks-within-the-network, which we train along with the overarching network. This \textit{meta-learning}~\cite{hospedales2020} generates potent neuron activation function sets, suggestive of orthogonal spanning functions, that increase the expressiveness and accuracy of the network.  

After discussing related work and our motivation, we describe how meta-learning diverse activation functions can generate better neural networks, as measured by difficult classification and nonlinear regression tasks. We show that learned diversity can enhance conventional neural networks as well as physics-informed neural networks, so the latter are doubly enhanced. To provide further insight into the advantages of diverse neuronal activations, we employ neuron participation ratios as a metric to elucidate the superior potential of these layers compared to their homogeneous counterparts. 
Additionally, we study the geometric nature of optimizing minima by examining the spectra of their Hessian matrices, shedding light on the underlying loss landscape of diversified neural networks.
Finally, by examining the interplay between stochastic processes and diversified neural networks, we gain valuable insights about how the synergy between the inherent randomness of the optimization procedure and learned diversity results in more generalizable models. We end by discussing future work and the potential for \textit{learned diversity} to enhance artificial neural networks, deep learning, and our appreciation of diversity itself.

\section*{Related Work} \label{sec:RelatedWork}

Researchers have recently begun to relax the rigid rules that have guided the development and use of artificial neural networks. Manessi and Rozza~\cite{manessi2018learning} investigate learning combinations of known neuronal activation functions, and Agostinelli et al.~\cite{Agostinelli2014} learn piecewise linear activation functions for each neuron. Apicella et al.~\cite{Apicella2020} survey trainable activation functions. Lau and Lim~\cite{Lau2018} review adaptive activation function in deep neural networks. Jagtap, Kawaguchi, and Karniadakis~\cite{JAGTAP2020} and Haoxiang and Smys~\cite{Haoxiang2021} include scalable hyper-parameters in their activation functions to improve their networks, while Qian et al.~\cite{QIAN2018204} linearly, nonlinearly, and hierarchically combine basic activation functions to optimize performance.

More radically, Gjorgjieva, Drion, and Marder~\cite{computational_div} investigate the computational implications of biophysical diversity and multiple timescales in neurons and synapses for circuit performance. Doty et al.~\cite{heterogeneous_cell_types} show that \emph{hand-crafted} heterogeneous cell types can improve the performance of deep neural networks. Xie, Liang, and Song~\cite{weight_div} demonstrate that diversity in synaptic weights lead to better generalization in neural networks. Mariet and Sra~\cite{diversity_dpp} sample a diverse subset of neurons and merge them with the remaining ones via a re-weighting procedure. Siouda et al.~\cite{Siouda2022} use genetic algorithms to optimize the number, forms, and types of hidden neurons. Hospedales et al.~\cite{hospedales2020} survey the current meta-learning landscape. Lin, Chen, and Yan~\cite{lin2014network} suggest nesting neural networks inside neural networks. 

Decisively, Beniaguev, Segev, and London~\cite{Beniaguev613141} write, “We call for the replacement of the deep network technology to make it closer to how the brain works by replacing each simple unit in the deep network today with a unit that represents a neuron, which is already -- on its own -- deep”, which is what we achieve here with our neuronal sub-networks that meta-learn sets of diverse activation functions that can outperform the corresponding homogeneous neural networks.


\begin{figure}[b!]
    \centering
    \vspace{-0.5cm}
    \includegraphics[width=0.7\linewidth]{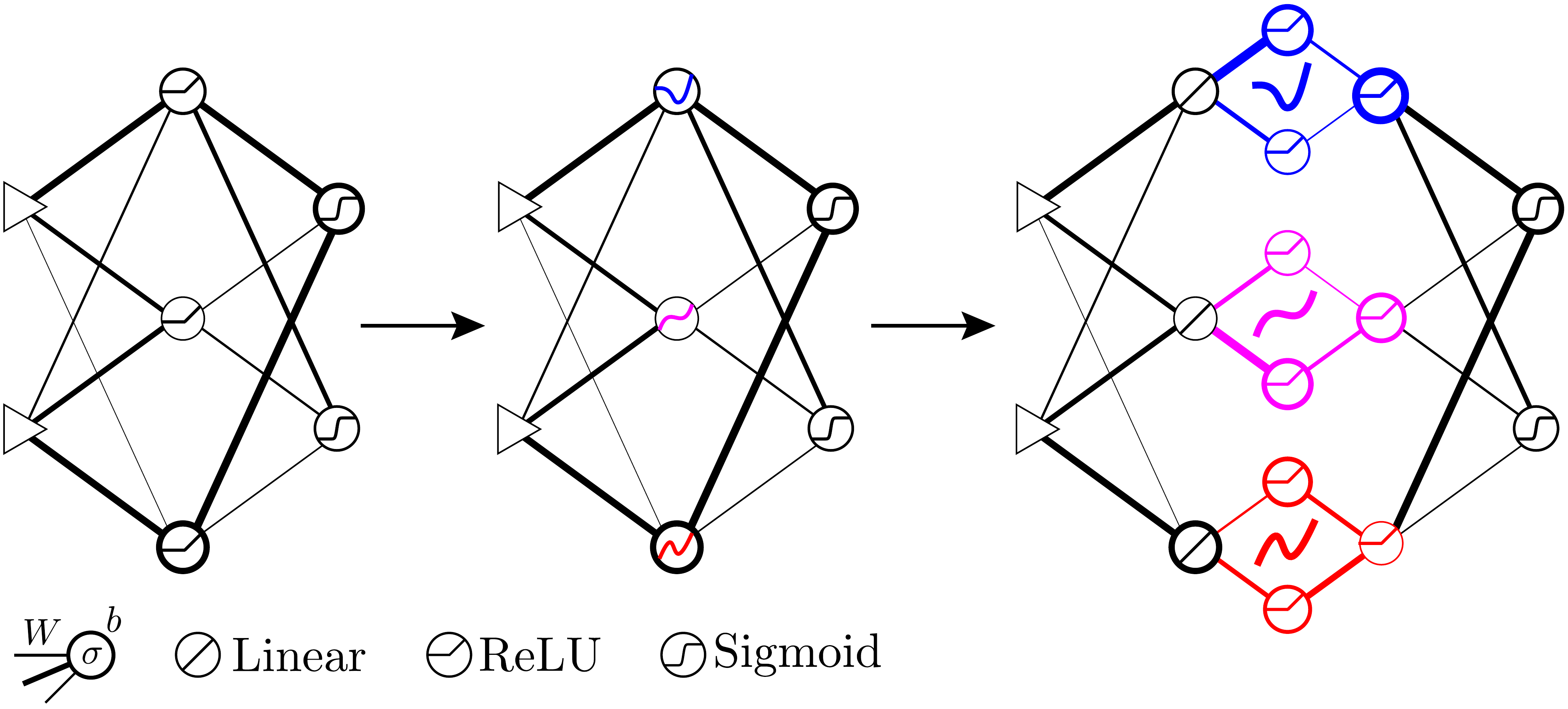}
   \caption{Progression from conventional artificial neural network to diverse neural network to learned diverse neural network. Line thicknesses represent weights $W$, circle thicknesses represent biases $b$, and sketches inside circles represent activation functions $\sigma$. 
   }
    \label{fig:div_schematic}
\end{figure}

\section*{Motivation} \label{sec:motivation}

Inspired by natural brains, feed-forward neural networks are nested nonlinear functions of linear combinations of activities
\begin{equation}
    a^\prime \stackrel{\text{vec}}{=} \sigma({\bf W}a+b),
\end{equation}
where the activation $\sigma$ is typically a saturating or rectifying function, and training strengthens or weakens the weights and biases $\mathbf{W}$ and $b$ to minimize an objective function, often called a ``cost'' or ``loss'' (from financial optimization).

Motivated by the well-studied mammalian visual cortex, varying neuronal activation functions by layer is common. However, within each layer, the activations are typically identical, as in Fig.~\ref{fig:div_schematic}~(left). Neural networks are universal function approximators~\cite{Cybenko1989,Hornik1991} and are often used to model hypersurfaces, either for classification or nonlinear regression. Varying the activations within a layer, as in Fig.~\ref{fig:div_schematic}~(middle), should therefore increase the expressiveness of the network by providing diverse spanning basis functions. Furthermore, replacing the activations by sub-networks, as in Fig.~\ref{fig:div_schematic}~(right), and training them for optimal results should increase the expressiveness even further. The training of the activation sub-networks can be on a different schedule than the training of the network, and the activations so obtained can be extracted from the sub-networks as interpolated functions and efficiently reused in other networks addressing different problems.

\section*{Algorithm} \label{sec:method}

\begin{figure}[t!]
    \centering
    \vspace{-2.0cm}
    \includegraphics[width=0.32\linewidth]{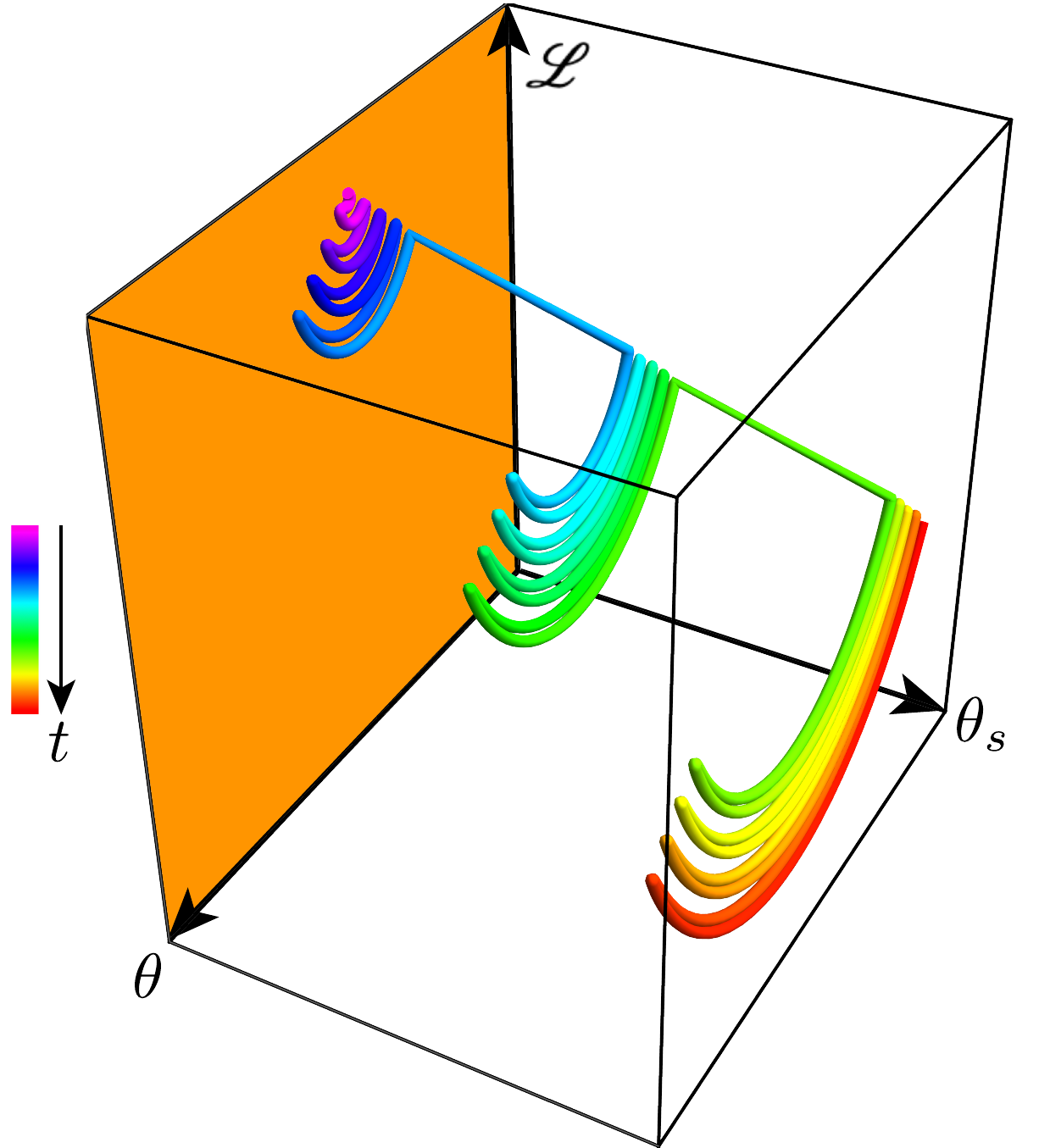}
    \vspace{-0.4cm}
    \caption{Schematic stochastic gradient descent meta-learning nested loops. Neural-network weights and biases $\theta$ adjust to lower losses $\mathcal{L}(\theta,\theta_s)$, during an inner loop, while periodically the \textit{sub-network} weights $\theta_s$ open extra dimensions and themselves adjust to allow even lower losses, during an outer loop. Rainbow colors code time $t$.} 
    \label{fig:MetaGradDescent}
\end{figure}

To create a learned diversity neural network (LDNN), incorporate sub-networks initialized to simple activations (like identity, ramp, or sigmoid functions). Train the network with many input-output pairs. Quantify the difference between the actual and expected outputs with a loss function $\mathcal{L}$. In an inner loop, compute the gradient of the loss function with respect to the network's weights and biases, and lower the loss by shifting its weights and biases down this gradient. In an outer loop, compute the gradient of the loss function with respect to the \textit{sub-networks'} weights and biases~\cite{maclaurin2015gradientbased}, and further lower the loss by shifting the sub-networks' weights and biases down \textit{this} gradient, thereby evolving new activations. Repeat to minimize loss. 

In the inner loop, the randomly shuffled inputs are the stochastic driver that buffets the network weights and biases $\theta$ as they adjust to lower the loss. In the outer loop, the activation sub-network weights and biases $\theta_s$  open extra dimensions or degrees of freedom to further lower the loss. Figure~\ref{fig:MetaGradDescent} provides an overview, and Algorithm~\ref{metalearn2} provides details.

\vspace{0.0cm}
\begin{algorithm}[H] \label{metalearn2}
\DontPrintSemicolon
\caption{Meta-learning activation functions $\sigma_n(\bullet)$ as sub-networks of network $f(\bullet)$, where $x \in X$ are training inputs, $ y \in Y$ are training outputs, $\hat y$ are network outputs, $R$ are learning rates, $\mathcal{L}$ are losses, $N$ are number of iterations, $N_T$ is number of neuron types, and $\theta = \{\mathbf{W},b\}$ are weights and biases. Subscript $I$ indicates the inner loop, which updates the network, and subscript $O$ indicates the outer loop, which updates the sub-networks. Network weights and biases update $N_I |X|$ times in the inner loop, while sub-network weights and biases update $N_{O}$ times in the outer loop.}
\SetKwFunction{LDNN}{LDNN}
\SetKwProg{Procedure}{Procedure}{}{}
\Procedure{\LDNN{$X, Y,\innerR, \outerR, \innerN, \outerN, N_T$}}{
    Initialize activation-function sub-networks $\varSigma = \{\sigma_n(\bullet ;\theta_s)\,\big|\, n = 1, \dots, N_T$ \}\;
    \For{$1, \dots, \outerN$}{
        $t \gets 1$\;
        Initialize network $f(x; \theta, \varSigma)$ with sub-networks $\varSigma$\;
        \For{$1, \dots, \innerN$}{          
            \ForAll{$(x, y) \in (X, Y)$}{
                Compute network prediction $\hat{y} \gets f(x; \theta, \varSigma)$\;
                Compute network loss $\mathcal{L}_t \gets \mathcal{L}(y, \hat{y})$\;
                Update network parameters $\theta \gets \theta - \innerR \nabla_{\theta} \mathcal{L}_t$ via gradient descent\;
                $t \gets t+1$\;
            }
        }
        Compute mean network loss $\mathcal{L}_{I} \gets \langle \mathcal{L}_t \rangle = \sum_t \mathcal{L}_t / N_I |X|$\;
        Update sub-network parameters $\theta_{s} \gets \theta_{s} - \outerR \nabla_{\theta_{s}} \mathcal{L}_{I}$ via gradient descent\;
    }
}
\end{algorithm}


\vspace{-0.25cm}
\section*{Results} \label{sec:results}

\subsection*{MNIST-1D} 

Here we implement~\cite{github} learned diversity neural networks with one hidden layer of 100 neurons and a cross-entropy loss function to classify the MNIST-1D data set, a minimalist variation of the classic Modified National Institute of Standard and Technology digits~\cite{deng2012mnist,greydanus2020scaling}. Each neuron type in the hidden layer is further instantiated by a feed-forward neural network of 50 hidden units evolved from a base sinusoid. We obtain similar results for different numbers of layers, different number of neurons per layer, and different base functions.

\begin{figure}[p]
    \centering
    \includegraphics[width=0.95\linewidth]{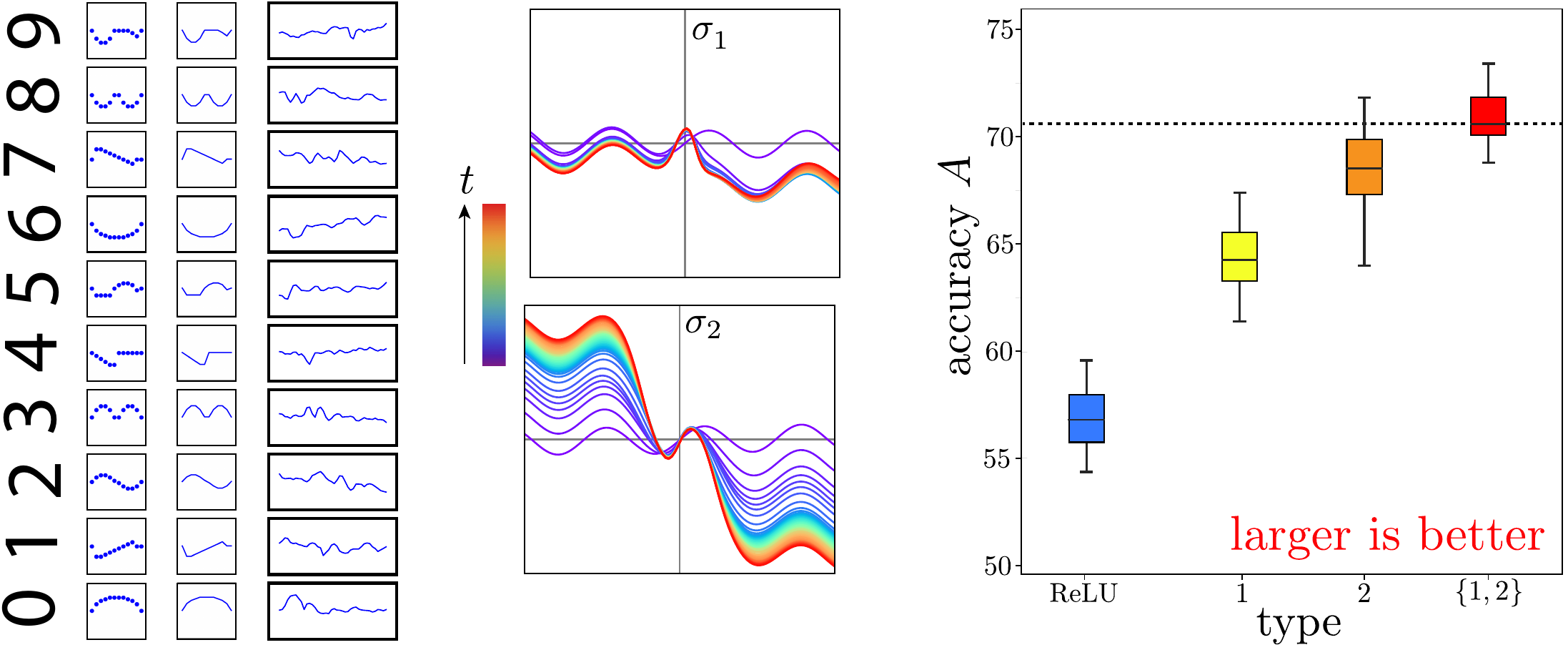}
    \caption{Meta-learning 2 activations for MNIST-1D classification. \textbf{Left:} Example MNIST-1D digit construction, rotated $90^\circ$ to emphasize the one-dimensionality of the digits. \textbf{Center:} Activation functions $\sigma_n(a)$ evolve from a base sinusoid, with violet-to-red rainbow colors encoding time $t$. \textbf{Right:} Box and whisker plots summarize distribution (including median, quartiles, and extent) of validation accuracy $A$ for a fully connected neural networks of 100 ReLU neurons (blue), type-1 neurons (yellow), type-2 neurons (orange), and a mix of type 1 and type 2 neurons (red). The mix of 2 neuron types outperforms any single neuron type on average.}
    \label{fig:Metalearning2}

    \vspace{1cm}
    \includegraphics[width=1.0\linewidth]{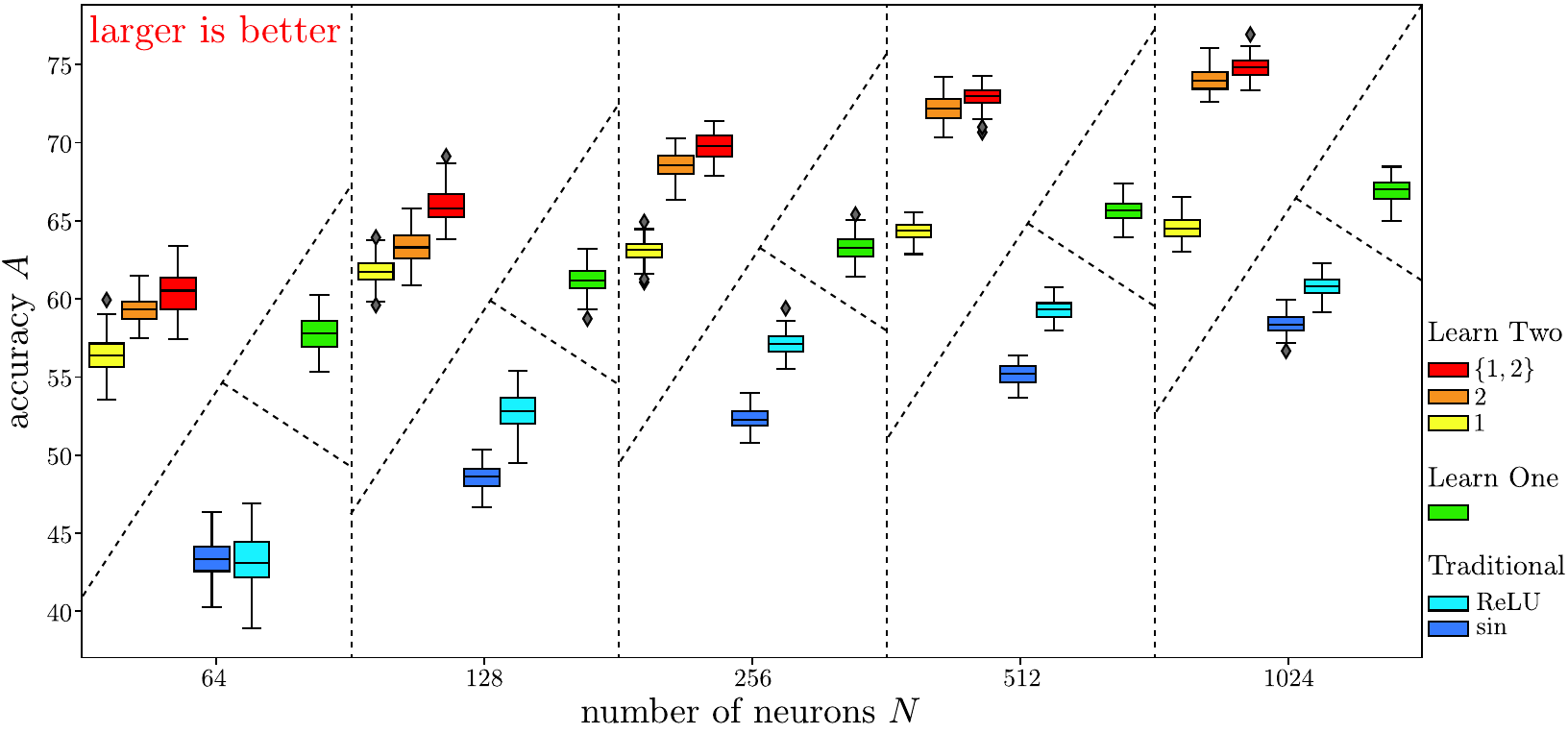}
    \caption{Neural network MNIST-1D classification accuracy as a function of network size. Box plots summarize accuracy distribution (including median, quartiles, extent, and outliers) for 100 initializations. Learning rate is optimized to avoid over-fitting but is the same for all network sizes. Activation functions evolved from zero (the null function) with similar results evolved from sine. Mixed networks of 2 neuron types outperform pure networks on average for all sizes and outperform both single learned activation and traditional activations. } 
    \label{fig:grid}
\end{figure}

Figure~\ref{fig:Metalearning2} summarizes meta-learning the activation functions of neurons in the hidden layer subject to the constraint of having two functions distributed equally among the neuronal population. Figure~\ref{fig:Metalearning2} (left) shows the construction of typical one-dimensional digits. Figure~\ref{fig:Metalearning2} (center)  show the evolution of the two activation functions, with time encoded as rainbow colors from violet to red. Figure~\ref{fig:Metalearning2} (right) shows box plots demonstrating validation accuracy for 50 fully connected neural networks composed of entirely $N_1$ type neurons (yellow), entirely $N_2$ type neurons (orange), and mixed type with $N_1$ and $N_2$ distributed equally among hidden layer (red). With the same training, the mixed network outperforms either pure network on average. These results are robust with respect to network size, as summarized by Fig.~\ref{fig:grid}.

\subsection*{van der Pol} 
\begin{figure}[ht!]
    \centering
    \includegraphics[width=0.95\linewidth]{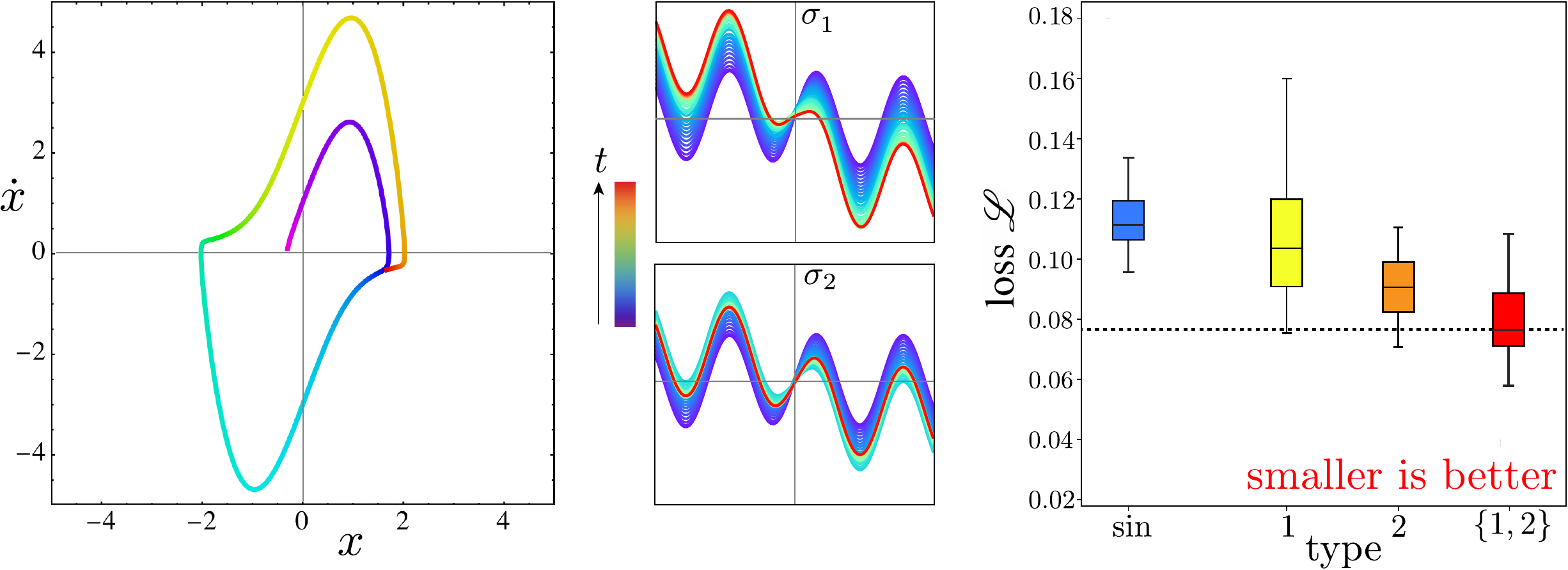}
    \caption{Meta-learning 2 activations for nonlinear regressing or  forecasting the van der Pol oscillator. \textbf{Left:} Typical orbit is attracted to a limit cycle, where rainbow colors code time $t$. \textbf{Center:} Activation functions $\sigma_n(a)$ evolve from a base sinusoid. \textbf{Right:} Box plots summarize distribution of neural network mean-square-error validation loss $\mathcal{L}$, starting from 50 random initializations of weights and biases, for a fully connected neural networks of sine neurons (blue), type-1 neurons (yellow), type-2 neurons (orange), and a mix of type 1 and type 2 neurons (red). The mix of 2 neuron types outperforms any single neuron type on average.} 
    \label{fig:vanDerPol}
\end{figure}

We obtain similar results for other tasks, such as nonlinear regression of the van der Pol oscillator~\cite{vanderpol}, which includes a linear restoring force and a nonlinear viscosity modeled by the differential equation
\begin{equation}
    \ddot x - \mu(1 - x^2)\dot x + x = 0,
\end{equation}
where the overdots indicate time derivatives. The van der Pol oscillator can model vacuum tubes and heartbeats and was generalized by FitzHugh~\cite{Fitzhugh} and Nagumo~\cite{Nagumo} to model spiky neurons. For viscosity parameter $\mu=2.7$, we trained neural networks to forecast the phase space orbit of the oscillator, as summarized by Fig.~\ref{fig:vanDerPol}. On average the learned diversity neural network outperforms either of its pure components as well as a homogeneous network of neurons with sinusoidal activations.

\subsection*{H\'enon-Heiles} 

The paradigmatic H\'enon-Heiles Hamiltonian~\cite{henon} 
\begin{equation}
   H=\frac{1}{2}\left(p_{x}^{2}+p_{y}^{2}\right) + \frac{1}{2}\left(x^{2}+y^{2}\right) + \left(x^2 y-\frac{1}{3}y^3\right)
\end{equation}
can model a star moving in a galaxy of other stars according to the Hamiltonian flow
%
\begin{equation} \label{hamEq}
    \left\{  \dot q,  \dot p  \right\}
    = \left\{ +\frac{\partial H}{\partial {p}}, -\frac{\partial H}{\partial {q}} \right\},
\end{equation}
where $q = \{x,y\}$ and $p = \{p_x, p_y\}$. Bounded motion is possible in a triangular region of position space. As orbital energy increases, circular symmetry degenerates to triangular symmetry, and integrable motion complexifies to chaotic motion.

\begin{figure}[p]
    \centering
    \includegraphics[width=0.8\linewidth]{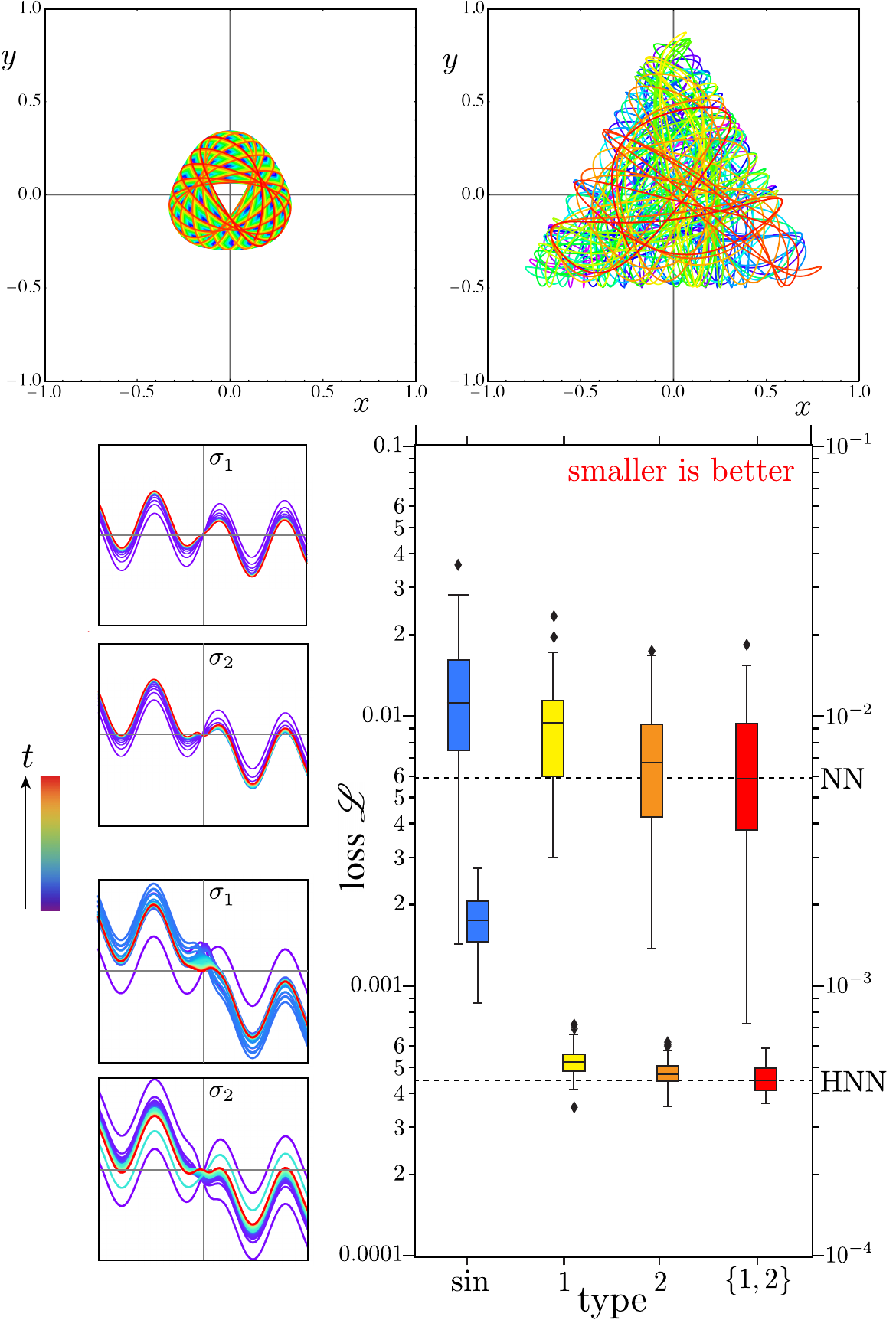}
    \caption{Meta-learning 2 activations for nonlinear regressing or forecasting H\'enon-Heiles orbits. \textbf{Top:} Regular and chaotic, low and high-energy H\'enon-Heiles orbits, where rainbow colors code time. \textbf{Bottom Left:} Conventional and Hamiltonian neural networks learn activation functions from base sinusoids. \textbf{Bottom Right:} Box plots summarize distributions of mean-\allowbreak square-\allowbreak error validation losses $\mathcal{L}$, starting from 50 random initializations of weights and biases, for fully connected neural networks. Hamiltonian neural networks greatly outperform conventional neural networks and heterogeneous neuron types consistently outperform their homogeneous components on average.}
    \label{fig:HNN}
\end{figure}

Consequently, for this example, we meta-learn activation functions for both a conventional and a Hamiltonian neural network~\cite{HNN,HGN, Choudhary2020,Miller,Miller2,Choudhary2}. Unlike conventional neural networks, which learn dynamical systems by intaking position and velocity and outputting their derivatives, a Hamiltonian neural network learns a dynamical system by intaking position and momentum and outputting a single energy-like variable, which it differentiates according to Hamilton's recipe. Rather than learning the derivatives, it learns the Hamiltonian function, which is the {\em generator} of derivatives. This more powerful and efficient strategy is an excellent example of physics-informed machine learning. 

More specifically, during training a conventional neural network (NN) maps positions and velocities $\{q_t, \dot q_t\}$ to approximations of their time derivatives, and adjusts its internal parameters to minimize the mean-square-error or loss
\begin{equation} \label{NNLossEq}
	\mathcal{L}_{\text{NN}} = \bigg\langle (\dot{q}_t-\dot{q})^2 + (\ddot{q}_t-\ddot{q})^2  \bigg\rangle_t.
\end{equation}
The trained network can extrapolate a given initial condition via the Euler update $ \{q, \dot q\} \leftarrow \{q, \dot q\} + \{\dot q, \ddot q\} dt$. By contrast, during training a Hamiltonian neural network (HNN) maps position and momenta $\{q_t, p_t\}$ to the scalar Hamiltonian function $H$, uses reverse-mode automatic differentiation to find the Hamiltonian's gradients, uses the gradients to approximate the position and momentum change rates, and adjusts its internal parameters to minimize the loss 
%
%
\begin{equation}\label{HNNLossEq}
    \mathcal{L}_{\text{HNN}}
    = \left\langle \left(\dot q_t - \frac{\partial H}{\partial p} \right)^2
    + \left(\dot p_t + \frac{\partial H}{\partial q} \right)^2 \right\rangle_t 
\end{equation}
and enforce Hamilton's motion equations. The trained network can extrapolate a given initial condition via the Euler update $ \{q, p\} \leftarrow \{q, p\} + \{\dot q, \dot p\} dt$.

As summarized by Fig.~\ref{fig:HNN}, the mix of 2 neuron types outperforms any single
neuron type on average for both conventional and Hamiltonian neural networks, but the Hamiltonian neural network is much better, and its mixed version is doubly enhanced. (Spread in Hamiltonian validation losses is much smaller than the spread in the conventional validation losses, possibly because enforcing symplectic structure on the loss manifold for the Hamiltonian neural network is a regularization that facilitates more consistent optimization, while the unbounded loss of the conventional neural network suffers greater variance due to the wide range of stable and chaotic trajectories.)


\subsection*{Pendulum Clock from Video} 
\begin{figure}[ht!]
    \centering
    \includegraphics[width=0.95\linewidth]{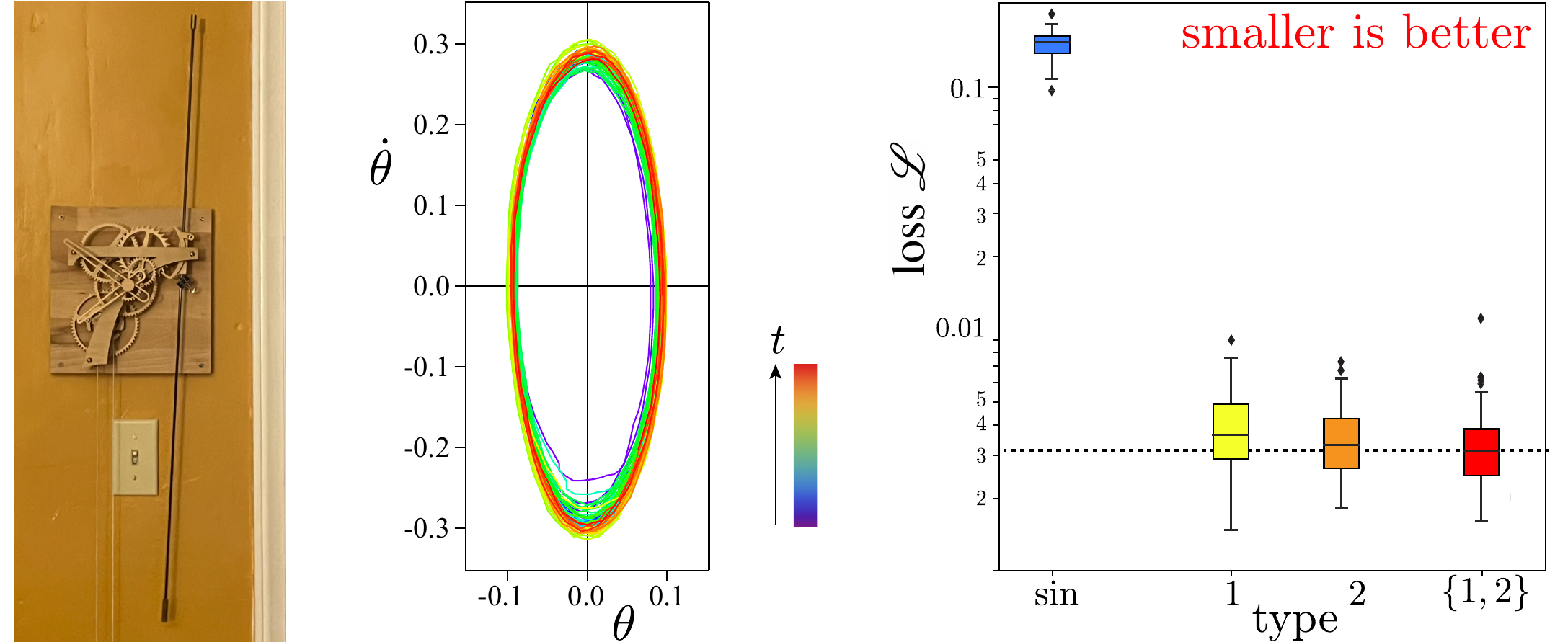}
    \caption{Meta-learning 2 activations for forecasting a real pendulum clock engineered to be almost Hamiltonian. \textbf{Left:} Falling weight (not shown) drives a wall-hanging pendulum clock. \textbf{Center:} State space flow from video data is nearly elliptical. \textbf{Right:} Box plots summarize distribution of neural network mean-square-error validation loss $\mathcal{L}$, starting from 50 random initializations of weights and biases, for a fully connected neural networks of sine neurons (blue), type-1 neurons (yellow), type-2 neurons (orange), and a mix of type 1 and type 2 neurons (red). Meta-learning diversity is a winning strategy.} 
    \label{fig:pendulumClock}
\end{figure}

As a final real-world example, we video recorded a wall-hanging pendulum clock, tracked the ends of its compound pendulum, and extracted its angles and angular velocities at equally spaced times~\cite{Choudhary2}. Engineered to be nearly Hamiltonian, the pendulum's Graham escapement periodically interrupts the fall of its weight as gravity compensates dissipation. We trained Hamiltonian neural networks to forecast its phase space orbit, as summarized by Fig.~\ref{fig:pendulumClock}. Once again, meta-learning proves advantageous.

\section*{Analysis} \label{sec:analysis}

To understand how mixed activation functions outperform homogeneous neuronal populations, we estimate the change in the dimensionality of the network activations. Start by constructing a neuronal activity data matrix {\bf X} with $N$ rows corresponding to $N$ neurons in the hidden layer and $M$ columns representing inputs. Each matrix element ${\bf X}_{ij}$ represents the activity of the $i^{th}$ neuron at the $j^{th}$ input. Center the activity so $\langle \bf X \rangle = 0$. Construct the neural co-variance matrix ${\bf C} = M^{-1}{\bf XX}^{T}$, which indicates how pairs of neurons vary with respect to each other, and compute the participation ratio
\begin{equation}
    \mathcal{R} 
    = \frac{(\operatorname{tr}{\bf C})^2}{\operatorname{tr}{\bf C}^2}
    = \frac{\left(\sum_{n=1}^N\lambda_n \right)^2}{\sum_{n=1}^N \lambda_n^2},
\end{equation}
where $\lambda_n$ are the co-variance matrix eigenvalues. If all the variance is in one dimension, say $\lambda_n = \delta_{n1}$, then $\mathcal{R} = 1$; if the variance is evenly distributed across all dimensions, so $\lambda_n = \lambda_1$, then $\mathcal{R} = N$. Typically, $1 < \mathcal{R} < N$, and $\mathcal{R}$ corresponds to the number of dimensions needed to explain most of the variance~\cite{Gao214262}. The normalized participation ratio $r = \mathcal{R} / N$.

\begin{figure}[th!]
    \centering
    \includegraphics[width=0.45\linewidth]{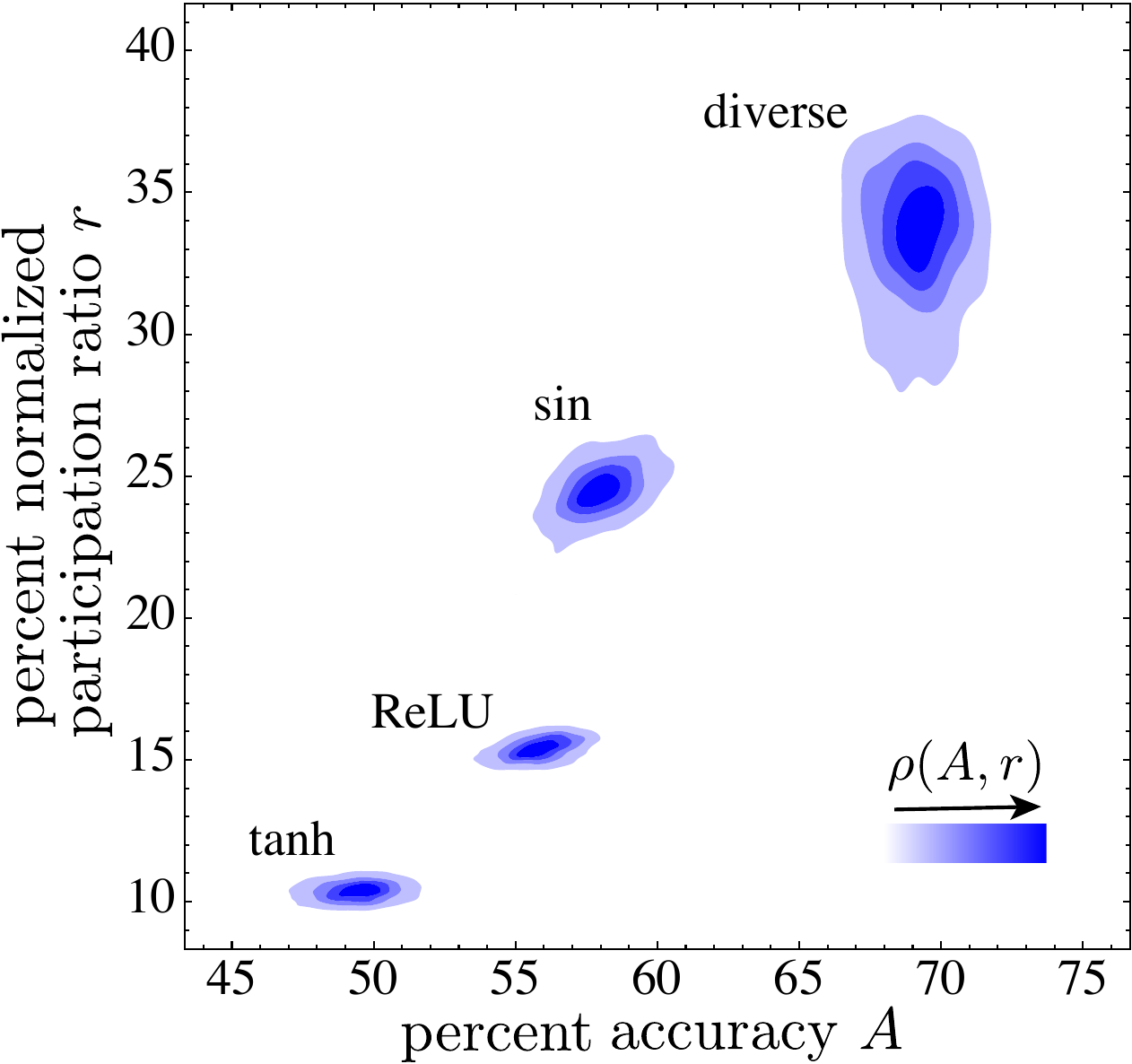}
    \caption{Probability densities $\rho(A,r)$ versus accuracy $A$ and normalized participation ratio $r = \mathcal{R}/N$ for multiple realizations of the Fig.~\ref{fig:Metalearning2} MNIST-1D heterogeneous network and three homogeneous networks with popular activation functions hyperbolic tangent, Rectified Linear Unit $f(x) = \max(0,x)$, and sine. Increased participation accompanies increased accuracy, with the diverse network maximizing both.}
    \label{fig:PR}
\end{figure}

\begin{figure}[ht!]
    \centering
    \includegraphics[width=0.7 \linewidth]{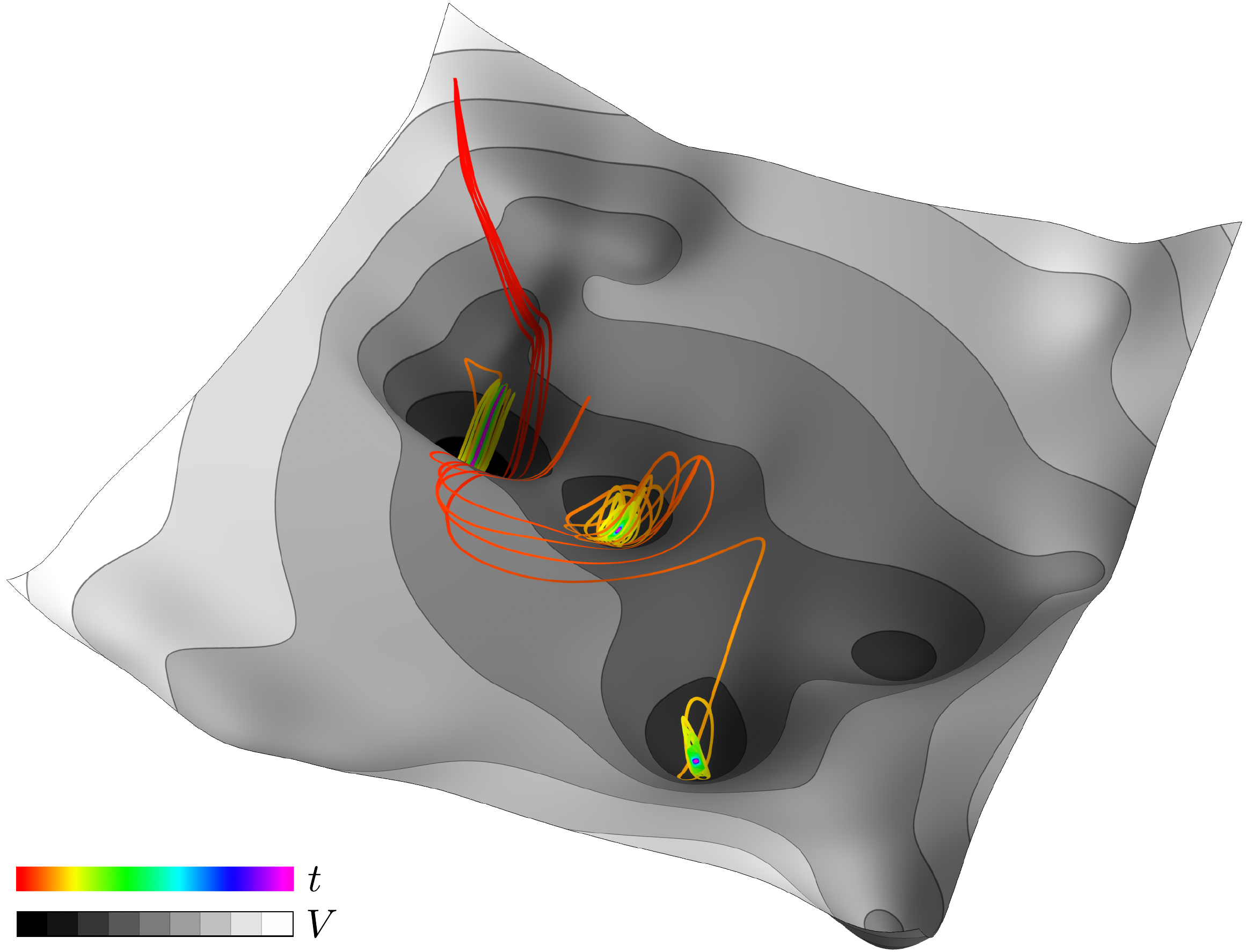}
    \caption{Noisy descent. Rainbow colors code time $t$ as state point wanders to different local minima of potential landscape $V$ from same initial conditions under multiple realizations of the same noise. } 
    \label{fig:SGD2}
\end{figure}
Figure~\ref{fig:PR} plots the joint probability densities $\rho(A,r)$ for multiple realizations of the Fig.~\ref{fig:Metalearning2} MNIST-1D learned diversity neural network and homogeneous competitors. The mix of two neurons types has the best mean accuracy $A$ and normalized participation ratio $r$, suggesting that more of its neurons are participating when the mix achieves the best MNIST-1D classification. In contrast, homogeneous networks of neurons with popular activation functions have lower accuracy and participation ratios reflecting their poorer effectiveness.

To understand the impact of learned diversity on the geometric nature of loss-function minima, we compute the spectrum of the Hessian matrix $\mathbf{H}=\nabla^{2}\mathcal{L}$, which captures the curvature of the loss function. Since $\mathbf{H}$ is a symmetric matrix, all its eigenvalues are real. A purely convex loss function would have a positive semi-definite Hessian everywhere. However, in practice, the loss function is almost always non-convex (with multiple local minima) due to the presence of hidden neuron permutation symmetries~\cite{perm_symmetries}. Therefore, understanding how diversity helps training find deeper minima is crucial.

Previous work suggests that flatter minima generalizes better to the unseen data~\cite{ghorbani2019,sankar2020}. For the Fig.~\ref{fig:Metalearning2} neural network meta-learning two neuronal activation functions, we find that once training has converged, the resulting minima from diverse neurons is flatter than from homogeneous ones, as measured by both the trace $\Tr\mathbf{H}$ of the Hessian and the fraction $f$ of its eigenvalues near zero: $\Tr \mathbf{H}_1 >  \Tr \mathbf{H}_2 > \Tr \mathbf{H}_{12}$ and $f_1 < f_2 < f_{12}$. If steep minima are harder for gradient descent to locate, then the flatter minima engineered and discovered by learned diversity neural networks imply enhanced optimization.


Stochastic processes can provide additional insights. Optimizing a neural network by randomly shuffling training data is like a noisy descent to a minimum in a potential landscape, as in Fig.~\ref{fig:SGD2}. The landscape is the network's cost or loss as a function of its weights and biases, and its shape depends on the neuron activation functions. The effective dynamics is that of an overdamped particle buffeted by noise sliding on a complicated potential with many local minima. The Langevin equation
\begin{equation}\label{eq:sdg_langevin}
    d\theta_{t} = - \nabla \mathcal{L}(\theta_{t})\, dt + \sqrt{2\mathbf{D}} \cdot d\mathcal{W}_{t}
\end{equation}
with noise intensity $\mathbf{D} = (\eta / B) \mathcal{L}(\theta) \mathbf{H}(\theta^*) $ describes the evolution of the weights and biases $\theta =\{W_{ij},b_i\}$ in a valley with local minimum $\theta^{*}$, where $\eta$ is the learning rate and $B$ is the training batch size~\cite{sdg_langevin_log,mandt2017stochastic,sirignano2019stochastic, chaudhari2019entropy}. The drift term with $dt$ includes minus the gradient of the loss function $\mathcal{L}$, and the Brownian motion noise term with $d\mathcal{W}_t$ includes the learning rate $\eta$. The noise aligns with the Hessian near a minimum, and the Eq.~\ref{eq:sdg_langevin} Hessian dependence ensures that stochastic gradient descent escapes multiple sharp minima via directions corresponding to large Hessian eigenvalues and eventually converges to a flatter minimum.

\section*{Conclusions} \label{sec:discussion}

Biomimetic engineering or biomimicry is design inspired by nature. Just as monoculture crops can be fragile, while diverse crops can be robust~\cite{Wetzel2016}, heterogeneous neural networks can outperform homogeneous ones. Here, we highlight advantages of varying activation functions \textit{within} each layer and learning the best variation by replacing activations by sub-networks.

Conceptually, learned diversity neural networks discover novel \textit{sets} of activation functions, when most artificial neural networks use just one of a small number of conventional activations per layer. Practically, mixes of learned activations can outperform traditional activations -- where even a $1\%$ improvement can be significant -- and the learned activations can be efficiently reused in diverse neural networks. Additionally, learned diversity can even improve already enhanced physics-informed neural networks like Hamiltonian neural networks~\cite{Tailin2019, Choudhary2020}. Future work includes optimizing learned diversity by adjusting hyperparameters, applying learned diversity to a wider range of  regression and classification problems, testing the diverse neural networks for robustness~\cite{robustness}, investigating clustering of learned activations, and applying learned diversity to different neural network architectures, such as recurrent neural networks and reservoir computers~\cite{rc_classic, optical_rc, quantum_rc}.

Learned diversity offers neural networks sets of tailored basis functions, which enhance their expressiveness and adaptability and facilitates efficient function approximation. \textit{When given the ability to learn their neuronal activation functions, neural networks discover heterogeneous arrangements of nonlinear neuronal activations that can outperform their homogeneous counterparts with the same training.} Our work provides specific examples of dynamical systems that spontaneously select diversity over uniformity, and thereby furthers our understanding of diversity and its role in strengthening natural and artificial systems. 


\section*{Methods}\label{sec:implementation}

We implement our neural networks in the Python programming language using the PyTorch open source machine learning library. We also implement them in the Python library JAX~\cite{jax2018github} using the JAX library Equinox~\cite{kidger2021equinox}. The code for the analysis and the network implementation can be found at our GitHub repository~\cite{github}.

Number of training pairs is  of order $10^4$, and number of training epochs is of order $10$. Due to computational constraints, the number of inner iterations is much smaller than the number of outer iterations. Indeed, the learner-meta-learner structure of the meta-learning algorithm incurs significant computational costs with a time complexity of $O(N_O N_I |X|)$. Current implementation of the algorithm is constrained by the number of inner loops within the outer loops since the inner loop is held in memory for the outer loop computation (such as the Algorithm~\ref{metalearn2} gradients $\nabla_{\theta_{s}} \mathcal{L}_{t}$) and optimization. In fact, this is one of the fundamental challenges of gradient-based meta-learning algorithms that currently limits the horizon of meta-optimization~\cite{short_term_horizon}. However, the inefficiency of the algorithm plausibly results from activation meta-learning being under-explored and ripe for improvement. 


PyHessian Library is used to compute hessian based statistics without the cost of generating the full hessian matrix. The trace of the hessian matrix is computed using Hutchinson's method exploiting the symmetric nature of the matrix~\cite{avron_randomized_2011}. The Empirical Spectral Density (ESD) of hessian eigenvalues is computed through Stochastic Lanczos Quadrature (SLQ)~\cite{ubaru_fast_2017} within several successive approximation schemes. Details can be found in Yao et al.~\cite{yao_pyhessian:_2020}. At an implementation level, a classifier or forecaster using the learned activation(s) is trained in Pytorch and the model is saved. Using this saved model and test data, PyHessian can use PyTorch's backward graph to compute the gradients needed to build the hessian trace and ESD.

The activation function is captured after meta-learning as the output of the learned activation networks on the interval $[-10,10]$ with 100 linearly spaced points. This output is then linearly interpolated between points and used as the activation function for the classifer at validation. Quadratic or cubic splines or symbolic regression can also be used. We need high order ($>10$) polynomials to fit the activation curves accurately so, while possible, we do not recommend polynomials as a reliable way to capture the features of the learned activation functions.



\bibliography{sample}

\section*{Acknowledgements}

This research was supported by O.N.R. grant N00014-16-1-3066 and a gift from United Therapeutics. W.L.D. thanks Kathleen Russell for the conceptualization of the original idea along with many subsequent discussions.

\section*{Author Contributions Statement}
A.C. designed and implemented our meta-learning code. A.R. trained and analyzed our neural networks and created our GitHub repository. J.F.L. lead the writing and finalized the figures. S.S. elucidated the diversity mechanism. W.L.D. motivated and guided the research. All authors contributed to the final manuscript.

\section*{Competing Interests}
The authors declare no competing interests.

\section*{Code Availability}
Our code is available at \url{https://github.com/nonlinearartificialintelligencelab/diversityNN}.

\end{document}